\documentclass[lettersize,journal]{IEEEtran}
\pdfoutput=1
\usepackage{amsmath,amsfonts}
\usepackage{algorithmic}
\usepackage{algorithm}
\usepackage{array}
\usepackage[caption=false,font=normalsize,labelfont=sf,textfont=sf]{subfig}
\usepackage{textcomp}
\usepackage{stfloats}
\usepackage{url}
\usepackage{verbatim}
\usepackage{graphicx}
\usepackage{cite}
\usepackage{CJKutf8}

\usepackage{float}
\usepackage{times}

\usepackage{soul}
\usepackage[hidelinks]{hyperref}
\usepackage[utf8]{inputenc}
\usepackage[small]{caption}
\usepackage{booktabs}
\usepackage{multirow}
\usepackage{multicol}
\usepackage{tabularx}
\usepackage{amssymb}
\begin{document}

\title{Benchmarking Continual Learning from Cognitive Perspectives}

\author{Xiaoqian Liu,
Junge Zhang,~\IEEEmembership{Member,~IEEE,}
        Mingyi Zhang,
        and Peipei Yang
\thanks{
This work is funded by the National Natural Science Foundation of China (Grand No. 61876181) and Beijing Nova Program of Science and Technology under Grand No. Z191100001119043 and in part by the Youth Innovation Promotion Association, CAS. (\textit{Corresponding author: Mingyi Zhang.})}
\thanks{Junge Zhang, Xiaoqian Liu, Mingyi Zhang and Peipei Yang are with the Institute of Automation, Chinese Academy of Sciences, Beijing, 100190, China (e-mail: jgzhang@nlpr.ia.ac.cn; xiaoqian.liu@ia.ac.cn; zhangmingyi2014@ia.ac.cn; ppyang@nlpr.ia.ac.cn.)}}


\maketitle

\begin{abstract}
Continual learning addresses the problem of continuously acquiring and transferring knowledge without catastrophic forgetting of old concepts. While humans achieve continual learning via diverse neurocognitive mechanisms, there is a mismatch between cognitive properties and evaluation methods of continual learning models.
First, the measurement of continual learning models mostly relies on evaluation metrics at a micro-level, which cannot characterize cognitive capacities of the model. Second, the measurement is method-specific, emphasizing model strengths in one aspect while obscuring potential weaknesses in other respects. 
To address these issues, we propose to integrate model cognitive capacities and evaluation metrics into a unified evaluation paradigm.
We first characterize model capacities via desiderata derived from cognitive properties supporting human continual learning.
The desiderata concern (1) adaptability in varying lengths of task sequence; (2) sensitivity to dynamic task variations; and (3) efficiency in memory usage and training time consumption. 
Then we design evaluation protocols for each desideratum to assess cognitive capacities of recent continual learning models. 
Experimental results show that no method we consider has satisfied all the desiderata and is still far away from realizing truly continual learning. Although some methods exhibit some degree of adaptability and efficiency, no method is able to identify task relationships when encountering dynamic task variations, or achieve a trade-off in learning similarities and differences between tasks.
Inspired by these results, we discuss possible factors that influence model performance in these desiderata and provide guidance for the improvement of continual learning models.

\end{abstract}



\section{Introduction}

\IEEEPARstart{M}{odern} machine learning has achieved remarkable success in a variety of artificial intelligence areas, such as computer vision\cite{DBLP:journals/corr/abs-2111-06377}, natural language processing\cite{DBLP:conf/naacl/DevlinCLT19,9380180}, robotics\cite{hayes2022online}, federated learning\cite{9477571}, etc. The powerful performance of existing machine learning models usually relies on large-scale offline training on independent and identically distributed (i.i.d.) data within one task. When new tasks are encountered, the models have to be retrained from scratch, resulting in waste of computational resources and loss of knowledge learned from previous tasks. By contrast, humans learn in a more flexible and efficient way that can continuously acquire new information and learn new skills throughout lifetime without catastrophic forgetting\cite{french1999catastrophic,kirkpatrick2017overcoming} of old memories. Therefore, an ideal machine learning model is expected to emulate human learning process and realize truly continual learning\cite{ring1994continual} to address dynamically changing environment.

\begin{figure}
\centering
\includegraphics[scale=0.13]{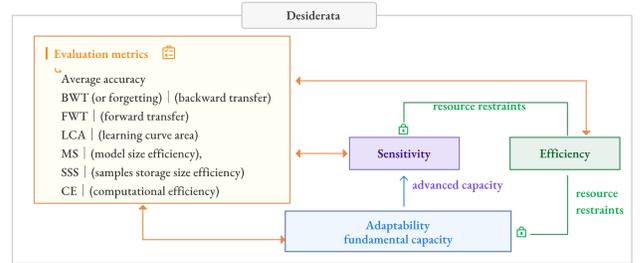}
\caption{Desiderata for continual learning. Adaptability is the most fundamental and core capacity to realize knowledge transfer and retention, while sensitivity is an advanced capacity to handle dynamic task variations. Adaptability and sensitivity are both restrained by efficiency in space and time. These desiderata can be measured by specific evaluation metrics designed for continual learning.}
\label{fig:desiderata}
\end{figure}


Recent findings in neuroscience and psychology have implicated some biological mechanisms that support continual learning in human brains. First, an appropriate balance between synaptic plasticity and consolidation underlies adaptive learning and long-term memory in human brains. Specifically, strengthening synaptic plasticity supports adaptive learning for new information\cite{gazzaniga2013cognitive} and can be realized by neurogenesis\cite{deng2010new}. Meanwhile, maintaining synaptic stability\cite{gazzaniga2013cognitive} through memory formation, storage and consolidation helps humans establish long-term memory, thus avoiding catastrophic forgetting of learned knowledge and skills. 
Second, humans are able to learn relationships between events or stimuli, either similarities or differences. These are encoded in memories via pattern separation and integration\cite{deng2010new,treves1992computational} to avoid interference between memories and retrieved via pattern completion\cite{rolls2006computational} to avoid catastrophic forgetting. 
Third, learning in human brains also requires to minimize extraneous cognitive load via information constraints so as to be efficient in processing and memorizing continuously emerging information.
When there is redundant and unnecessary information presented, the redundancy effect\cite{Jin2012} may occur, resulting in an extraneous cognitive load and thus hindering learning. 


As the human brain is a unified organism, the realization of continual learning by humans does not simply depend on a certain biological mechanism, but involves the interaction between various mechanisms that influence each other or promote and inhibit\cite{BARRON202185} mutually under different stimuli. Therefore, an ideal machine learning model for continual learning is also supposed to consider a variety of cognitive properties for learning and memory. In this paper, we propose desiderata for continual learning models based on cognitive properties that support human continual learning as discussed above. The desiderata include: (1) adaptability; (2) sensitivity; and (3) efficiency. Relationships among the desiderata with their relations to specific evaluation metrics are shown in Fig.~\ref{fig:desiderata}. Specifically, adaptability addresses knowledge transfer and retention between old and new tasks, and indicates the potential of a continual learning model in handling varying lengths of task sequence. Having achieved adaptability, a continual learning model is also required to be able to identify various relationships between tasks, which is an advanced model capacity in realizing continual learning and considered as sensitivity to dynamic task variations. Moreover, an ideal continual learning model is also supposed to be efficient in memory usage and training time consumption, which is the third desideratum we propose. Analysis of these desiderata with their relations to cognitive properties is introduced in Section~\ref{property}.

However, existing evaluation protocols for continual learning models have not focused on cognitive properties embedded in human brains, which neglects some important model aspects that should be examined and cannot assess whether current continual learning models have realized truly continual learning. 
Firstly, mainstream evaluation protocols only consider evaluation metrics such as average accuracy across tasks, forgetting or backward transfer (BWT)\cite{lopez2017gradient}, which only measure continual learning models at a micro-level but cannot characterize their cognitive capacities from a methodological perspective.
For example, it is assumed that a lower value of forgetting implies higher model capacity to retain old knowledge. Yet a continual learning model with extremely low average accuracy across tasks can also obtain a low forgetting value, as the model has learned little knowledge with nothing to be forgotten. As a result, it cannot offer a panorama of model cognitive capacities. 
Therefore, there demands for a systemic and comprehensive evaluation paradigm to contextualize the relation of model performance to human-like cognitive properties in realizing continual learning. 

Additionally, experimental set-ups in current evaluation protocols for continual learning models are usually method-specific, which emphasizes the strengths of a model in one aspect while obscuring possible weaknesses in other respects. For instance, Task-IL under multi-head settings\cite{Gido2019three} has been widely used in continual learning evaluation. However, such a setting hides the true difficulty of continual learning problem \cite{farquhar2018towards}. Therefore, a model-agnostic evaluation paradigm is needed for probing strengths and limitations of continual learning methods in terms of their cognitive capacities in different aspects.
Accordingly, we further propose a systemic evaluation paradigm consisting of evaluation protocols designed for each desideratum to assess model performance. 

To summarize, this paper aims to propose a systemic and comprehensive evaluation paradigm that can be used as a unified evaluation framework for continual learning models. The framework not only consists of specific evaluation metrics but also desiderata for continual learning methods derived from relevant neurocognitive theories. Thus, it is different from previous comparative studies on continual learning, which mostly compare existing continual learning methods from either a theoretical\cite{GGerman2019Continual,hadsell2020embracing,lesort2020continual,kudithipudi2022biological} or experimental perspective\cite{Matthias2021Defying,Marc2021Class,kemker2018measuring,pfulb2019comprehensive,farquhar2018towards,Eden2021Comprehensive,mundt2022clevacompass}. 

Results on experiments across eight benchmarks reveal that no method we consider has satisfied all the desiderata, despite that some methods exhibit some degree of adaptability and efficiency. Findings show that most methods evaluated fail in long task sequence and obtain insufficient adaptability on large and difficult datasets. Moreover, no method is able to identity various task relationships to address dynamic task variations. Particularly, regularization-based methods tend to focus on learning similarities between tasks but with the need of explicit task information to help infer task identity. Replay methods, instead, tend to focus on learning differences between tasks, and dynamic architectures seem to have no preference for types of task relationships. 
As for efficiency, there can be an optimal replay buffer size for replay methods that balances memory budgets and model performance given a certain task sequence. Meanwhile, most methods evaluated can be applied to online continual learning\cite{chaudhry2018efficient} without sacrificing much performance. The main contributions of this paper are as follows.
\begin{itemize}
    \item We propose desiderata derived from neurocognitive theories for machine learning models to realize truly continual learning on a basis of cognitive properties supporting continual learning in human brains.
    
    \item We propose a systemic and comprehensive evaluation paradigm composed of evaluation protocols designed for each desideratum to assess performance in adaptability, sensitivity and efficiency of recent continual learning methods from cognitive perspectives.
    
    \item Experimental results reveal that no method we consider has satisfied all the desiderata we proposed, and thus no existing machine learning models have achieved adequate cognitive capacities for continual learning. Especially, these methods cannot address dynamic task variations to achieve a trade-off between learning similarities and differences between tasks, despite that some methods exhibit some degree of adaptability and efficiency. With possible factors discussed, we look ahead to the future direction of continual learning and provide guidance for model improvement.
\end{itemize}

\section{Definition and Desiderata}

In this section, we first formalize the continual learning problem by addressing its dynamic learning characteristics in contrast to traditional machine learning. Then we propose three desiderata for continual learning models based on cognitive properties that support learning and memory in human brains. General principles of evaluation protocols designed for each desideratum are also introduced.

\subsection{Problem Formulation}
\label{problem formula}

Although there has not been a unified definition for continual learning in machine learning, we can generally formalize continual learning as in Eq.~\eqref{eq1} following\cite{lesort2020continual}.
Specifically, given a potentially infinite sequence of unknown data distributions $D=[D_1, D_2,...,D_N]$ from $N$ tasks, continual learning aims to learn an optimal prediction model $h_i$ on the current data distribution $D_i$ without catastrophically forgetting old knowledge learned from previous tasks on $[D_1, D_2,..., D_{i-1}]$. A task in continual learning is considered to correspond to a data distribution in the sequence $D$.

\begin{equation}
<^*h_i, M_i> \gets <^*h_{i-1}, ^*X_i, M_{i-1}, t_i>, \forall D_i \in D
\label{eq1}
\end{equation}
where the prediction model $h_i$ and the training set $X_i$ of task {\it i} are indispensable elements (denoted by an asterisk mark) in the problem formulation, while the replay buffer $M_i$ and task label $t_i$ are optional. The replay buffer stores data sampled from previous training sets and the task label contains explicit task information of each task to help distinguish task identity.


Based on above, we can formalize the continual learning problem as a dynamic learning process. Specifically, a prediction model $h_{i-1}$ trained on sequential tasks with training sets $[X_1,X_2,...,X_{i-1}]$ will change its parameter update direction towards a new optimization objective when adapting to a new task with training set $X_i$ as input. Theoretically, the change can be continuous as the model continues to encounter new data distributions from subsequent tasks. With the help of a replay buffer or task labels, the prediction model can leverage sample data from old tasks and distinguish task identities between two adjacent data distributions. However, neither the replay buffer nor the task label is always available for continual learning. This formulation of continual learning problems then poses challenges on machine learning models. It requires the model to be able to rapidly adapt to new tasks while remembering and leveraging knowledge learned from old tasks, which emulates the process of human learning and memory.

\subsection{Desiderata for Continual Learning Models}
\label{property}

In consistence with the problem formulation of continual learning, we propose three desiderata for continual learning models on a basis of cognitive properties that support lifelong learning in human brains: (1) adaptability; (2) sensitivity; and (3) efficiency.

\subsubsection{\textbf{Adaptability}}

Basically, continual learning requires a machine learning model to learn in an uncertain and changing environment. Likewise, learning and memory in human brains also involve dynamic changes in the effectiveness of neural connections\cite{kandel2013principles,bear2016neuroscience}. Learning enables humans to continuously acquire new information and rapidly adapt to new environments, while memory avoids catastrophic forgetting of old knowledge or skills learned from previous experiences. Learning can be realized by neurogenesis and recent evidence demonstrates that neurogenesis is not limited to early stages of human development but throughout adulthood\cite{gazzaniga2013cognitive,kandel2013principles}. However, an interruption during learning can make people forget, suggesting that memories are initially held in a fragile form\cite{bear2016neuroscience}. To survive interruption and memory loss, the brain selectively transfers short-term memory into long-term memory which is much more robust. 

To achieve continual learning that emulates learning and memory in human brains, a machine learning model needs to have strong adaptability to quickly transfer to new tasks while protecting old knowledge and skills from interference by new information, thus avoiding catastrophic forgetting. Then the adaptability of a continual learning model can be evaluated from two dimensions: (1) knowledge transfer, the capacity to rapidly adapt to new tasks, and (2) knowledge retention, the capacity to preserve old knowledge and skills.
Specifically, the capacity of knowledge transfer and retention can be measured by average accuracy across tasks combined with BWT. 
Furthermore, model performance in varying lengths of task sequence should be examined to assess adaptability. Details of evaluation protocols of adaptability can be referred to Section~\ref{scheme}.


\subsubsection{\textbf{Sensitivity}}

Continual learning concerns the change of tasks that determines relationships between tasks, which also affect model performance. Therefore, the second desideratum we propose for continual learning is to be sensitive to dynamic task variations. 
Typically, an ideal continual learning model is supposed to identify various task relationships and achieve a trade-off in learning similarities and differences between tasks.
Such a model capacity is similar to the pattern separation and integration in human brains. Pattern separation is an ability to distinguish two closely related images, episodes or spatial configurations that the storage and recall of explicit memory rely on\cite{kandel2013principles}. It is an essential step in information processing to avoid memory interference\cite{treves1992computational}. 
Pattern integration is the ability of immature dentate granule cells to provide an association between events and add similarity between representations of encoded memories, which is critical to the global pattern separation function of the dentate gyrus\cite{deng2010new}. 

Inspired by above, we evaluate the sensitivity to dynamic task variations through three different continual learning scenarios: Class-IL (Class-IL), Domain-IL (Domain-IL), and Task-IL (Task-IL)\cite{hsu2018re}.
These scenarios are differentiated by types of differences between tasks, involving the differences in input distributions, output distributions, and output spaces. Additionally, an ideal continual learning model needs to be capable of both fine-grained and coarse-grained tasks regardless of how task changes. Thus, we also evaluate the sensitivity at varying task granularity, which concerns the degree of similarities and differences between tasks. Specifically, tasks with more similar training data in semantics can be considered as a fine-grained task sequence, such as classifying different species belonging to the same genus, while a coarse-grained task sequence has training data in more different semantic meanings, such as discriminating cats from planes. Details of evaluation protocols of sensitivity are introduced in Section~\ref{scheme}.

\subsubsection{\textbf{Efficiency}}

One important advantage of continual learning over traditional machine learning is that it does not require retraining from scratch when dealing with new tasks. Therefore, continual learning is assumed to be more efficient. However, there are also resource constraints for continual learning models as the memory and computational budgets are always limited in practice.
In some cases, if there is additional unnecessary data provided for model training, it will not only lead to a waste of computational resources but may also decrease model performance, similar to the redundancy effect in cognitive load theory. The redundancy effect refers to the phenomenon where learning is hindered by the introduction of unnecessary interacting elements compared to the presentation of less information\cite{Jin2012}. Reasons are that the additional elements have to be processed in working memory, resulting in extraneous cognitive load\cite{plass2010cognitive}. 

We then propose efficiency as the third desideratum for continual learning models, which is measured with respect to memory usage and training time consumption. Specifically, we evaluate model performance with varying sizes of replay buffer and varying numbers of training epochs given a certain task sequence. A continual learning model then is supposed to make a balance between model performance, replay buffer size and training time consumption. Moreover, training data of each task in continual learning is supposed to be provided one-by-one in a single pass, requiring the model to be trained only using one epoch and thus learn tasks sequentially. This evaluation thus can also help figure out whether recent continual learning models are able to satisfy the single-pass training requirement. Details of evaluation protocols of efficiency are presented in Section~\ref{scheme}.

\begin{figure}
\centering
\includegraphics[scale=0.23]{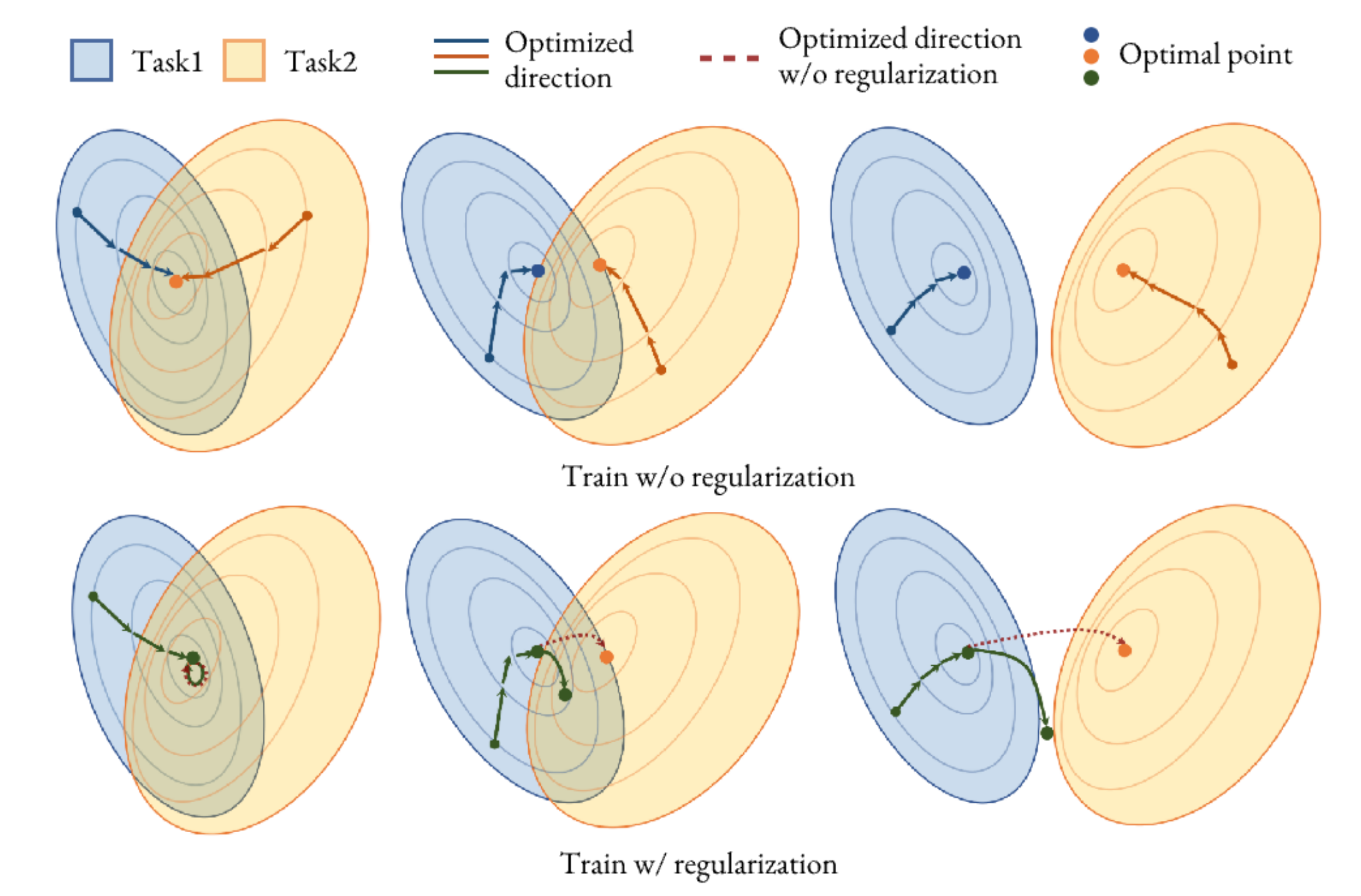}
\centering
\caption{
Regularization-based approaches for continual learning. \textbf{Top}: Training sequentially without regularization. After learning the first task, the model is optimized towards the area of low error for Task1 (blue area) to reach the optimal point for Task1 (blue point). When learning Task2, if gradients are updated only towards the area of low error for Task2 (yellow arrow), the model will deviate from the optimization space of Task1. Consequently, knowledge learned from Task1 will be catastrophically forgotten. 
\textbf{Bottom}: Training sequentially with regularization. The model is optimized towards the intersection (green arrow) of optimization spaces of each task and is supposed to reach the optimal point (green point) for task sequences.}
\label{fig:regular}
\end{figure}

\begin{figure*}[h]
	\centering
	\includegraphics[scale=0.17]{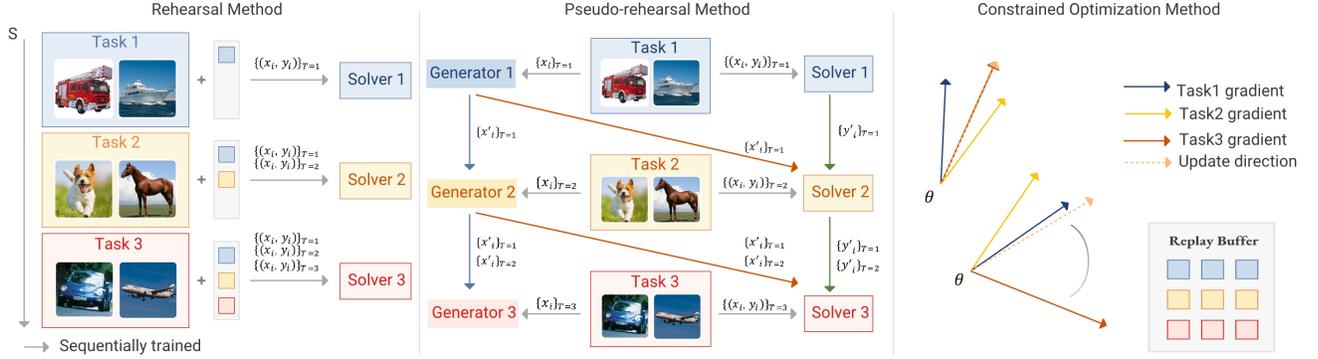}
	\caption{Replay approaches for continual learning.
\textbf{Left}: Rehearsal methods select and store sample data from old tasks and are retrained on old samples with data from new tasks. $(x_i, y_i)$ denotes sample training data and corresponding labels from each task.
\textbf{Middle}: Pseudo-rehearsal methods store pseudo samples (denoted by $(x^{\prime}_i)$ generated via a generative model (i.e., VAE or GAN) in replay buffer and these pseudo samples are learned with data from new tasks. The generative model (denoted by Generator) is optimized together with the pseudo-rehearsal model (denoted by Solver).
\textbf{Right}: Constrained optimization methods sample and store a subset of old data in replay buffer. The old samples are used to constrain the direction of gradient update when adapting to a new task by using relevant information of old samples\cite{pellegrini2019latent}.}
\label{fig_latent_replay}
\end{figure*}

\begin{figure}
\centering
\includegraphics[scale=0.2]{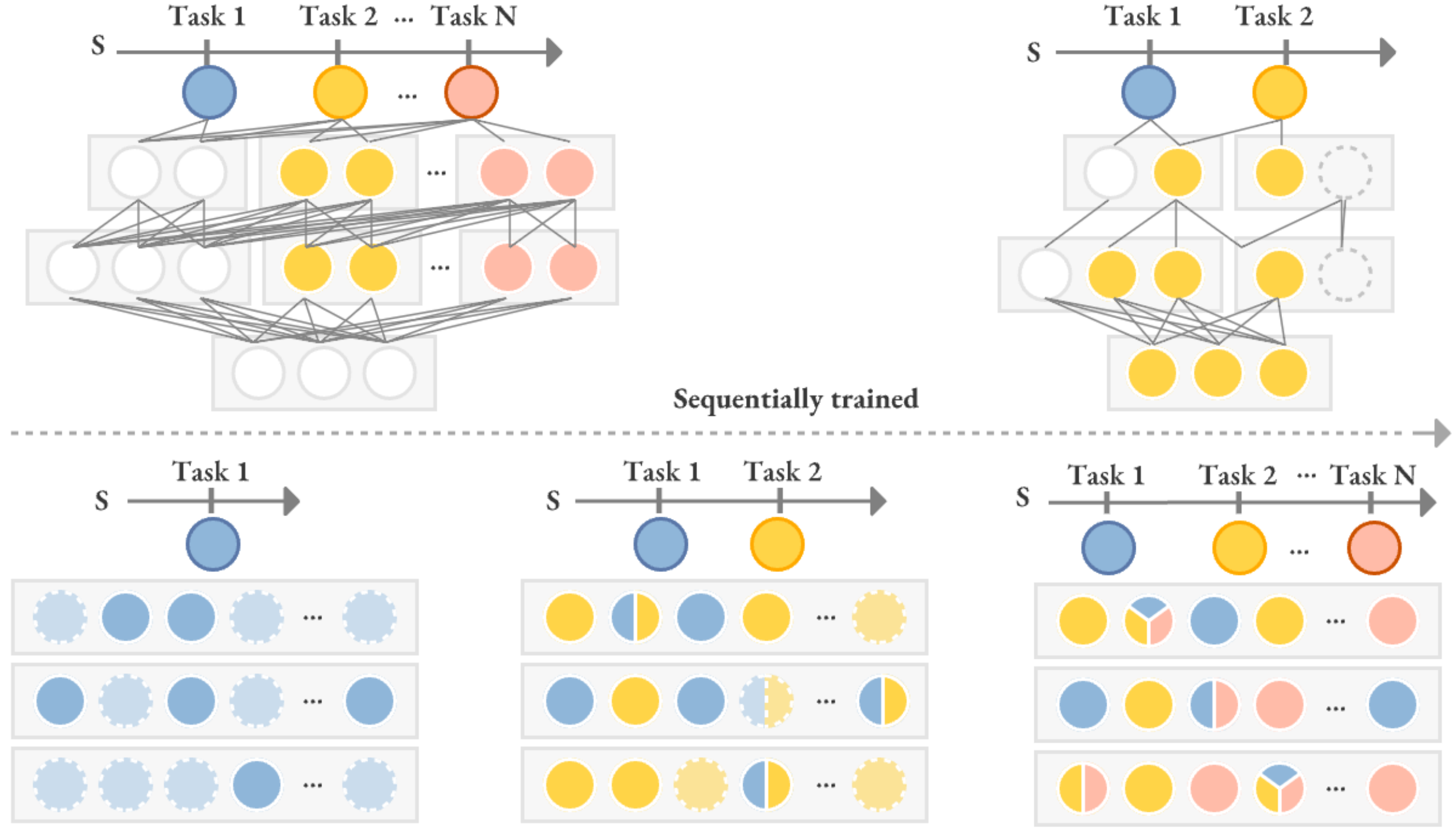}
\centering
\caption{Dynamic architectures for continual learning. \textbf{Top}: expandable dynamic architectures where new neurons or modules are developed for learning new tasks without interference with previous architectures, which have two variants. The first variant shown in \textbf{Left} retains old network architectures throughout training and adds new model capacity to reuse old computations and learn new ones simultaneously \cite{DBLP:journals/corr/RusuRDSKKPH16}.
The second variant in \textbf{Right} selectively
retrains old network and expands model capacity when necessary, thus dynamically deciding its optimal
capacity when learning new tasks\cite{DBLP:journals/corr/abs-1708-01547}.
\textbf{Bottom}: fixed dynamic architectures where neurons or modules are separated and dedicated to different tasks without interference between each other. Neurons in desaturated colors have not been activated when learning the current task and the ones in mixed colors have been activated and utilized in multiple tasks.}
\label{fig: dynamic}
\end{figure}

\section{Continual Learning Approaches}
The dynamic characteristics and desiderata of continual learning pose challenges on traditional machine learning models.
In this section, we review the mainstream methodology of existing studies on continual learning and categorize the methods into three families: (1) regularization-based approaches; (2) replay approaches; and (3) dynamic architectures. Although there is some overlapping with the categorization in \cite{Matthias2021Defying,Marc2021Class}, we aim to provide an analysis for existing works in terms of their solutions to catastrophic forgetting that are inspired by learning and memory in human brains.

\subsection{Regularization-based Approaches}
Catastrophic forgetting\cite{1989Catastrophic} in continual learning is mainly caused by different loss function surfaces between tasks. Specifically, as a back-propagated neural network adapts to a new task, parameters that have been learned on old tasks will be distorted, resulting in the loss of old knowledge. To address this issue, a straightforward way is to impose an explicit constraint on the loss function by adding a regularization term as a penalty. These are regularization-based approaches, the concept of which is shown in Fig.~\ref{fig:regular}. Such a solution to alleviate catastrophic forgetting is similar to synaptic plasticity and stability in human brains, where synaptic connections are selectively strengthened according to the function of synapses for learning and memorization. Likewise, regularization-based approaches\cite{lin2022trgp} promote or inhibit gradient updates according to the importance of parameters for old tasks.

Regularization-based approaches can be further divided into two categories: (1) approaches based on parameter importance \cite{kirkpatrick2017overcoming,Aljundi17,Zenke17,RN5,nguyen2017variational,NIPS2019_8690}, which impose restrictions on the changes of important parameters for previous tasks; and (2) approaches based on knowledge distillation\cite{Li17learning,castro2018end,Wu19Large,Hou_2019_CVPR,Liu_2020_CVPR,Zhao_2020_CVPR}, which integrate old knowledge into new task learning.
However, methods based on parameter importance are difficult to find appropriate indicators of parameter importance, and performance will degrade when handling long-sequence tasks. Additionally, methods based on knowledge distillation usually demand for a small sample set of old tasks and the issue of imbalance between old and new classes can also result in ineffective learning.

\subsection{Replay Approaches}
\label{sec:replay methods}

Neuroscience studies have shown that hippocampal replay is a necessary component of memory consolidation\cite{carr2011hippocampal} and Complementary Learning System (CLS)\cite{mcclelland1998complementary} suggests that hippocampal replay can reduce catastrophic forgetting and improve generalization during human learning. Inspired by this, replay approaches for continual learning store raw samples of previous tasks in a replay buffer or generate pseudo-samples using a generative model. When learning a new task, old examples are replayed from the replay buffer or the generative model to alleviate catastrophic forgetting.

According to the way of collecting and selecting old samples, replay approaches can be categorized into three families as shown in Fig.~\ref{fig_latent_replay}, including (1) rehearsal approaches\cite{rebuffi2017icarl,riemer2018learning,aljundi2019online,buzzega2020dark}; (2) pseudo-rehearsal approaches\cite{shin2017continual,riemer2019scalable,pellegrini2019latent,pourkeshavarzi2022looking}; and (3) constrained optimization approaches\cite{lopez2017gradient,chaudhry2018efficient,aljundi2019gradient,farajtabar2020orthogonal}. Typically, rehearsal approaches are simple and effective by retraining old raw samples to preserve old knowledge. However, different sampling strategies have a large impact on performance. Pseudo-rehearsal approaches do not require old samples and are appropriate for scenarios with high data privacy requirements. Nonetheless, they rely on an additional generative model which also needs to address catastrophic forgetting. Constrained optimization approaches can achieve positive backward transfer and interfere with old tasks as minimal as possible but usually with low learning efficiency and high computational cost.

\subsection{Dynamic Architectures}

To explicitly model the dynamics of continual learning, another line of continual learning methods aims to establish dynamic architectures to alleviate catastrophic forgetting. Inspired by neurogenesis and functional specialization of brain regions, the dynamic architectures dedicate different network parts or model parameters to each task and dynamically adjust the architecture for learning new tasks, while freezing previous task parameters or masking out network parts for learning old tasks. The mechanism of dynamic architectures for continual learning is shown in Fig.~\ref{fig: dynamic}.

Dynamic architectures\cite{rusu2016progressive,aljundi2017expert,yoon2017lifelong,DBLP:journals/corr/FernandoBBZHRPW17,Ju2021Reinforced,mallya2018packnet,serra2018overcoming} have achieved some success on continual learning based on adaptively optimized framework, network quantization and pruning, but still inferior to other continual learning methods in terms of training efficiency and model complexity. In addition, due to the high computational cost during training, the development of dynamic architectures has been still restricted with shallow networks. However, from the perspective of neuroscience, dynamic architectures are a promising solution to continual learning as it characterizes the dynamics of neuron activities in human continual learning.
Then future work that aims to improve the training efficiency of dynamic architectures with low model complexity can further boost this research direction.

\begin{table*}[t]
\caption{Dataset statistics. ``Num. of tasks'' denotes the length of task sequence in continual learning. There are 9 tasks in the New Classes (NC) scenario and 8 tasks in the New Instances (NI) scenario from CORe50. ``Num. classes per task'' denotes the number of classes to be distinguished in each task with respect to corresponding number of tasks in ``Num. of tasks''.}
\label{tab:dataset}
    \centering
    \begin{tabular}{l|c|c|c|c|c}
    \toprule
    Dataset & Num.of tasks & Input size
     & Num. classes per task  & Num. training images & Num. test images \\
     \hline
    SplitMNIST & [1,2,5]& 1$\times28\times28$
    & [10,5,2] & 60000 in total & 10000 in total \\
    PermutedMNIST  & 5  & 1$\times28\times28$
    & 10 & 60000 in total & 10000 in total \\
    RotatedMNIST & 5  & 1$\times28\times28$
    & 10 & 60000 in total & 10000 in total \\
    SplitCIFAR10 & [1,2,5]  & 3$\times32\times32$ & [10,5,2]
    & 5000 per class & 1000 per class \\
    SplitCIFAR100 & [1,2,4,5,10,20,25 50] 
    & 3$\times32\times32$
    & [100,50,25,20,10,5,4,2] & 500 per class & 100 per class\\
    SplitTinyImageNet & [5,10,20,50] & 3$\times64\times64$ & [40,20,10,4] & 500 per class  & 50 per class \\
    CORe50 & 9/8 & 3$\times32\times32$
    & 10 at category/50 at object level
    & 120000 in total  & 45000 in total \\
    ``ImageNet50'' & 9 & 3$\times32\times32$
    & 10 at category/50 at object level 
    & 120000 in total  & 45000 in total  \\
    \bottomrule
    \end{tabular}
\end{table*}

\section{Experiments}
We conduct comprehensive and in-depth comparative experiments on seven representative continual learning methods with fine-tuning (Naive) as a baseline across eight qualitatively different datasets. In this section, we introduce the methods in comparison, datasets, evaluation metrics, and details of evaluation protocols designed for each desideratum.

\subsection{Methods in Comparison}
Learning without Forgetting (\textbf{LwF}) \cite{Li17learning} uses distillation techniques to integrate the knowledge of network trained on old tasks into the learning process for new tasks. Elastic Weight Consolidation (\textbf{EWC})\cite{kirkpatrick2017overcoming} uses the second derivative of loss function to measure the importance of each parameter so as to update parameters in a targeted manner. Synaptic Intelligence (\textbf{SI})\cite{mack2013principles} models parameters of each neuron as a vector and the path integral of parameter changes between old and new tasks is used as a regularization item to measure the importance of each parameter for the current task. 
Incremental Classifier and Representation Learning (\textbf{iCaRL}) \cite{rebuffi2017icarl} replays sample data stored in a memory buffer when learning a new task, and leverages the distillation technique to alleviate catastrophic forgetting. Averaged GEM (\textbf{A-GEM)}\cite{chaudhry2018efficient} ensures that the average loss of old samples in memory buffer does not increase as new tasks are encountered. Gradient based Sample Selection (\textbf{GSS})\cite{aljundi2019gradient} formulates sample selection as a constraint reduction problem by selecting a fixed subset of constraints that best approximate the feasible region defined by original constraints.
\textbf{HAT}\cite{serra2018overcoming} is a dynamic architecture which relies on a task-based hard attention mechanism to maintain the information learned from previous tasks with fixed neural parts allocated to each task learning.

\subsection{Datasets and Evaluation Metrics}
Datasets with statistics details can be referred to Table~\ref{tab:dataset}. Note that we construct ``ImageNet50'' as a variant of CORe50, the structure of which remains the same as CORe50. For evaluation metrics, we mainly adopt \textbf{Average Accuracy} and \textbf{Backward Transfer}\cite{lopez2017gradient}.

\subsection{Evaluation Protocols}
\label{scheme}

As simply adopting evaluation metrics cannot characterize cognitive capacities of continual learning methods, we propose desiderata derived from cognitive properties of human continual learning for model assessment from a methodological and panoramic view.
The aim of this evaluation paradigm is to contextualize the relation of model performance to cognitive capacities that emulate continual learning activities in human brains. 
This subsection introduces details of evaluation protocols for each desideratum proposed in Section~\ref{property}. Unless specified, the default learning environment is class incremental setting without the provision of task labels.

\textbf{Adaptability} Adaptability evaluation of continual learning methods aims to examine and compare model capacity of knowledge transfer and retention when adapting to new tasks. Experiments include training and testing continual learning models on SplitMNIST and SplitCIFAR10 involving varying numbers of tasks (1, 2 and 5 tasks), and on SplitCIFAR100 and SplitTinyImageNet with 5, 10, 20 and 50 tasks, respectively. When handling varying lengths of task sequence, model performance on SplitCIFAR100 is also examined with the number of tasks varying from 1, 2, 4, 5, 10, 20, 25 to 50 tasks.

\textbf{Sensitivity} Sensitivity evaluation of continual learning methods focuses on model capacity in identifying relationships between tasks. Specifically, we examine and compare model performance in three different learning environments. The experiments involving class incremental and Task-IL are conducted across four benchmarks including SplitMNIST, SplitCIFAR10, SplitCIFAR100 and SplitTinyImageNet, each with a task sequence consisting of 5 tasks. The Domain-IL experiments are conducted on two variations of MNIST dataset including PermutedMNIST and RotatedMNIST, each with a task sequence comprising 5 tasks.

To further scrutinize model capacity of identifying similarities and differences between tasks during continual learning, experiments involving varying task granularity are conducted on CORe50\cite{lomonaco2017core50}. Specifically, image classification performed at object level (50 classes) is considered as fine-grained tasks and classification at category level (10 classes) is regarded as coarse-grained tasks. 
Moreover, two different learning scenarios are considered: New Classes (NC) and New Instances (NI)\cite{lomonaco2017core50}. 
We also apply a new dataset, ``ImageNet50'', where the original 50 domestic objects belonging to 10 categories in CORe50 are replaced by data from ImageNet. Specifically, ``ImageNet50'' consists of 10 categories belonging to animals and each category contains five different species. Experiments on ``ImageNet50'' only considers the NC scenario.

\textbf{Efficiency} Efficiency evaluation of continual learning methods concerns memory usage and training time consumption by a model. To evaluate the efficiency in memory usage, we examine and compare the performance of three representative replay methods (iCaRL, A-GEM and GSS) with varying sizes of replay buffer in the range [100, 300, 500, 700, 1000, 2000, 5000]. The number of training epochs is set to 1 for all the tested methods across three benchmarks, each involving a task sequence consisting of 5 tasks. We also compare model performance in varying numbers of training epochs in the range [1, 5, 10, 20, 50, 100]. We do not directly calculating the training time of each model as it can be influenced by a variety of factors, including the hardware used, CUDA, experimental platform, etc. Thus, it is difficult and usually unfair to directly compute training time without strict and unified configurations.

\begin{figure*}[h]
\centering
\includegraphics[scale=0.9]{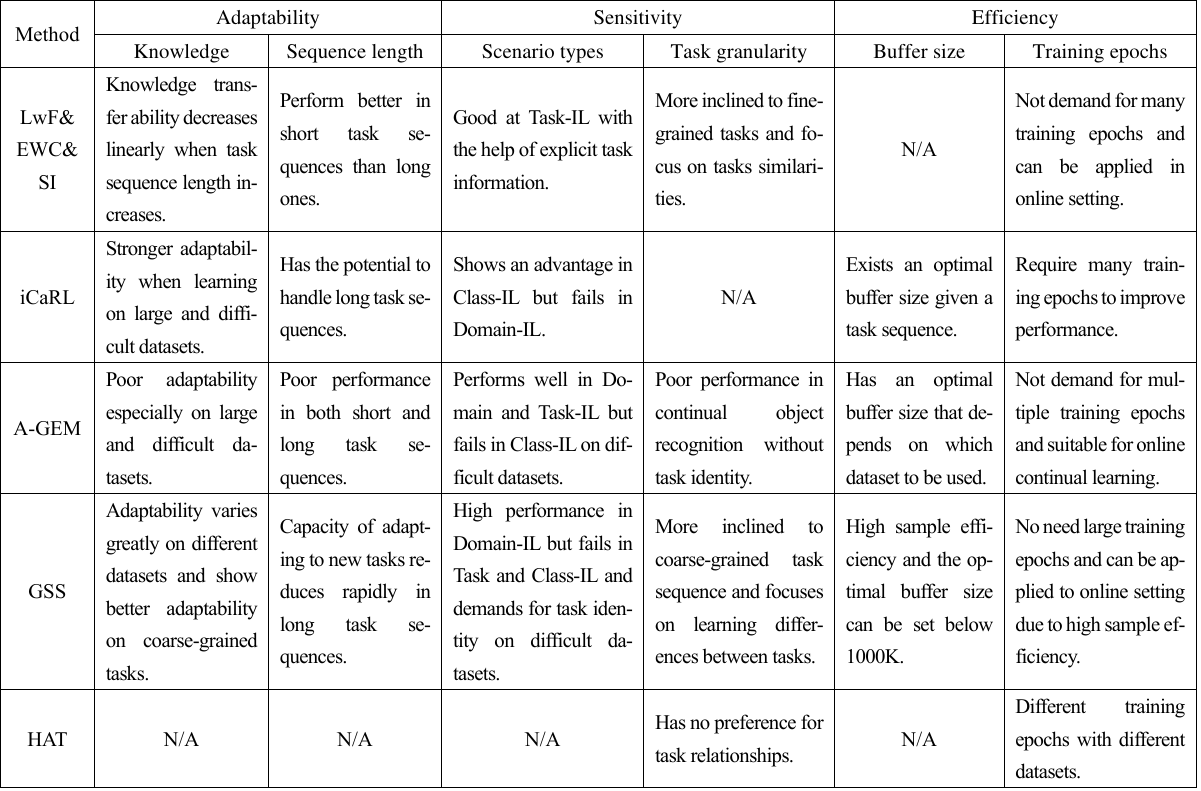}
\caption{Results of continual learning methods with respect to desiderata.}
\label{tab:result}
\end{figure*}

\section{Results and Discussion}
\label{results}
This section categorizes capacities of continual learning methods evaluated in the experiments by desiderata proposed in Section~\ref{property}. Strengths and weaknesses of these continual learning models are summarized concerning different desiderata. We also discuss factors that may influence the model performance in different aspects so as to provide insights for the improvement of continual learning methods.

\subsection{Overall Results}
Performance of methods for comparison in three desiderata is summarized in Table~\ref{tab:result}. 
Particularly, three regularization-based methods (LwF, EWC and SI) perform similarly in nearly all the experiments. In addition, we only describe the performance of HAT in experiments evaluating its sensitivity at different task granularity and efficiency in training time consumption. The reason is that HAT has to be evaluated in task incremental setting and thus cannot be equally compared to other methods whose default setting is Class-IL.


Experimental results show that no method we consider has satisfied all the desiderata for continual learning. There are some possible factors influencing model performance in these desiderata. For example, regularization-based methods usually fails in a long task sequence and tend to focus on learning similarities between old and new tasks, as their optimization objective is to find the intersection of target spaces of different tasks. Such methods optimize model parameters towards the region of intersection, thereby realizing knowledge transfer and retention. However, with the increase of task sequence length, the area of intersection gradually narrows and in most cases, it is of high probability that the intersection would be a null set, leading to significant performance degradation in adaptability. 

For replay methods, they can be significantly affected by the number of samples in each task, the number of classes to be distinguished and the differences between samples from each task. For instance, the gradient based sample selection in GSS helps to select representative samples from old tasks to maximize the diversity of data distribution stored in the replay buffer. However, the effect relies heavily on the sample differences between tasks as GSS calculates the cosine distance between feature vectors of sample data from adjacent tasks to measurement the sample difference. As a result, differences between samples and the accuracy of measuring the sample difference via cosine distance directly affect the performance of GSS. For A-GEM, there might be two reasons accounting for its poor performance in adaptability and sensitivity. First, A-GEM randomly samples data from old tasks and the samples can be of poor quality that cannot appropriately approximate the distribution of old data. Second, A-GEM calculates the mean of gradients on samples to represent the gradient of all old samples, which is prone to gradient offset. Consequently, the gradient mean may not direct to the ideal parameter update direction.

Moreover, there are some factors leading to higher performance of iCaRL in some cases especially in adaptability. For example, the nearest-mean-of-exemplars classifier in iCaRL is robust against changes in data distribution. Firstly, exemplars selected for learning the classifier are prioritized with a herding-based step. Additionally, representation learning in iCaRL relies on the use of exemplars combined with distillation to avoid catastrophic forgetting. The complicated network architecture of iCaRL also benefits its representation learning via effective feature extraction especially on large and difficult datasets. Importantly, the number of categories to be learned in each task has been already known by iCaRL during initialization. Therefore, during inference, iCaRL can effectively distinguish categories according to features learned from each category.

To conclude, each continual learning method examined in the experiments has its strengths and weaknesses in terms of each desideratum. No method has satisfied all the desiderata and factors influencing their performance vary in different aspects. Based on findings, we provide perspectives for model improvement in continual learning in Section~\ref{future}.

\begin{figure*}[t]
\centering
    \begin{minipage}{0.45\textwidth}
    \centering
    \includegraphics[scale=0.27]{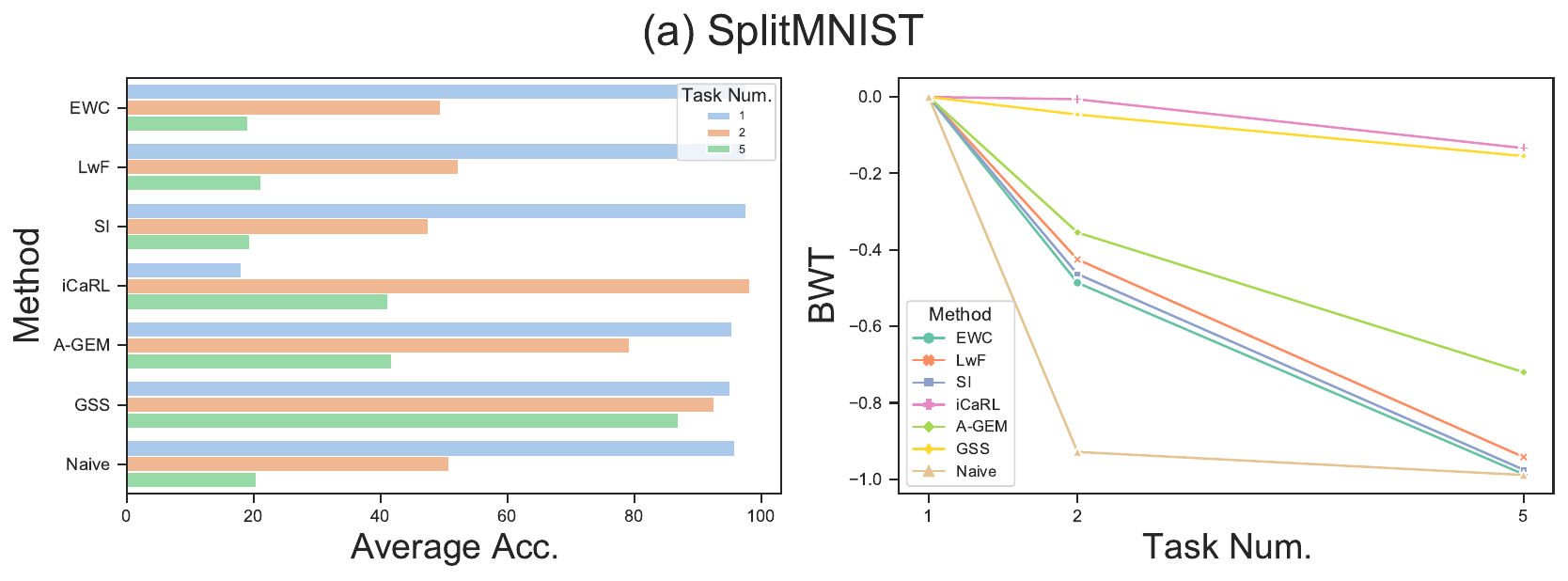}
    \end{minipage}
    \begin{minipage}{0.45\textwidth}
    \centering
    \includegraphics[scale=0.27]{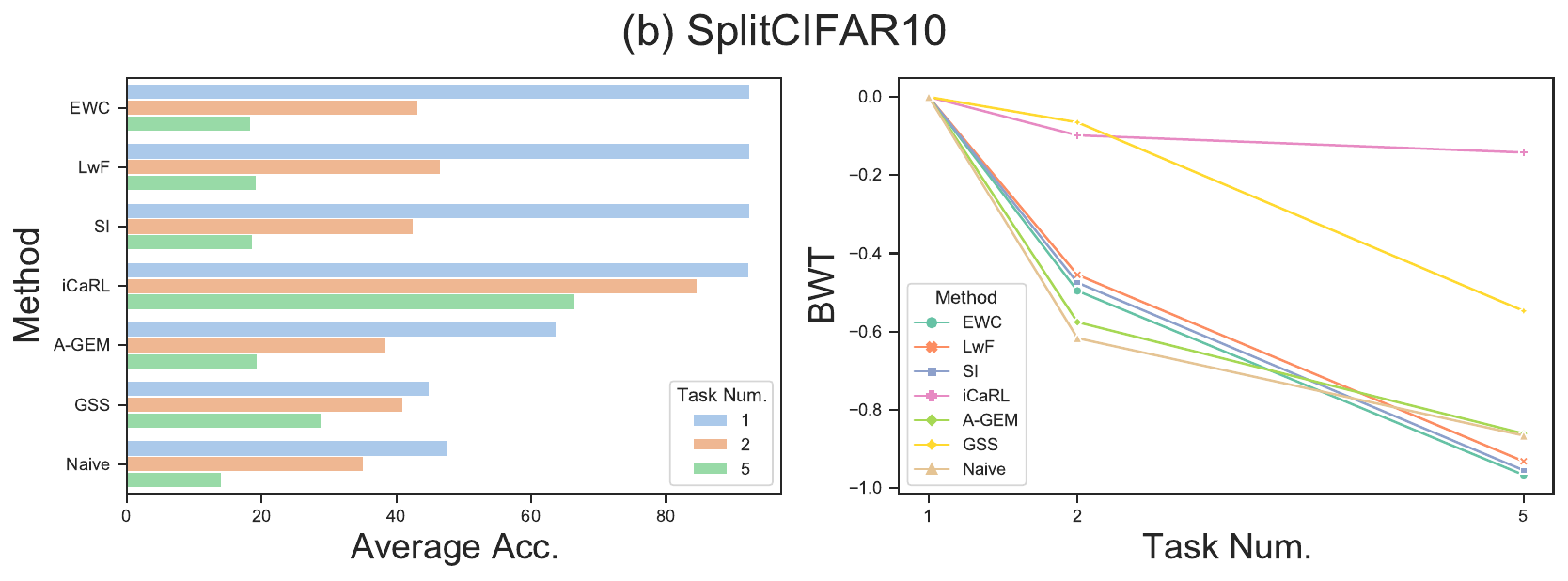}
    \end{minipage}
    \begin{minipage}{0.45\textwidth}
    \centering
    \includegraphics[scale=0.27]{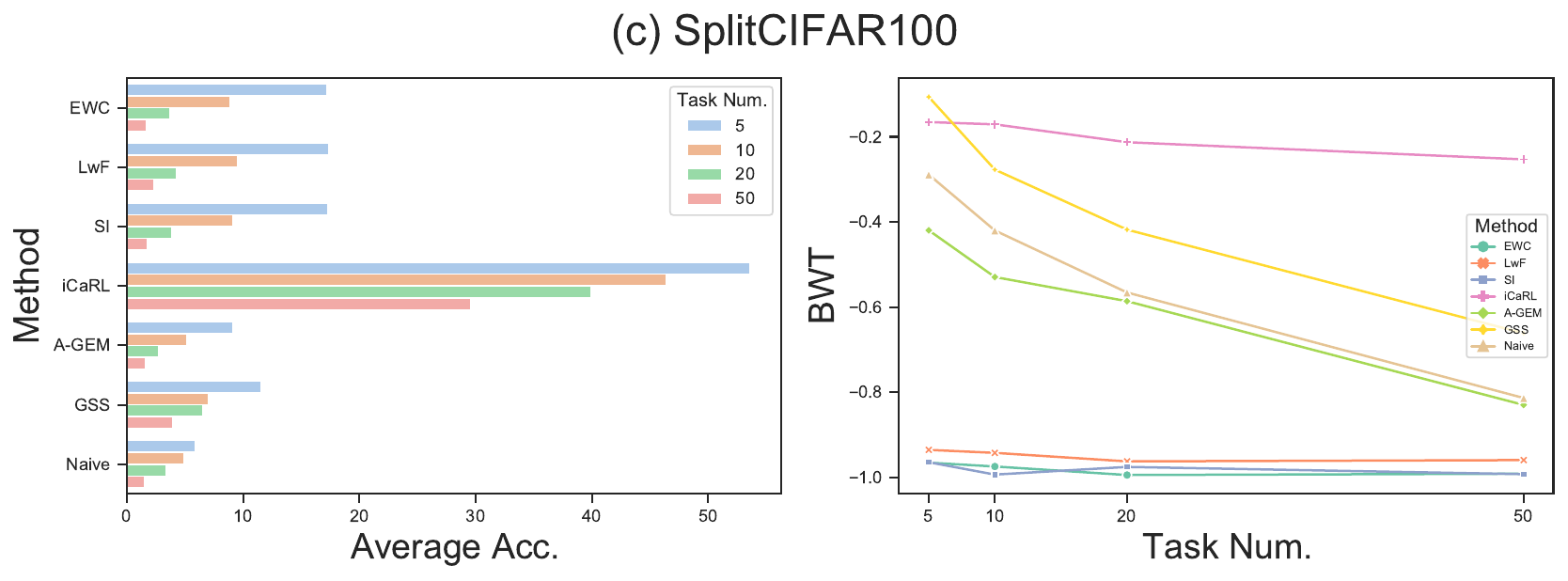}
    \end{minipage}
    \begin{minipage}{0.45\textwidth}
    \centering
    \includegraphics[scale=0.27]{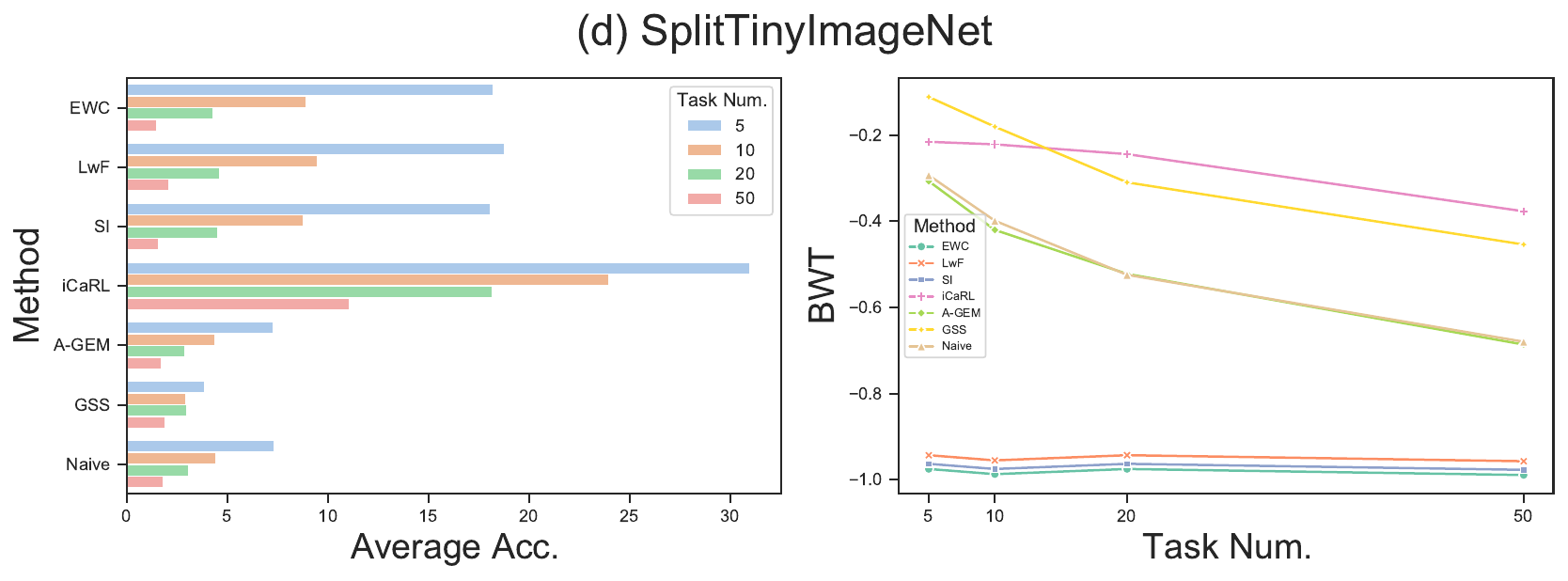}
    \end{minipage}
\caption{Adaptability results of knowledge transfer and retention across four benchmarks.  
All the methods evaluated exhibit a decreasing trend of average accuracy and backward transfer (BWT) as the number of tasks increases. However, iCaRL shows an advantage on SplitCIFAR100 and SplitTinyImageNet and has the potential to handle long task sequences. iCaRL and GSS achieve higher BWT across four datasets.}
\label{fig:adapt}
\end{figure*}

\begin{figure}
\centering
\includegraphics[scale=0.32]{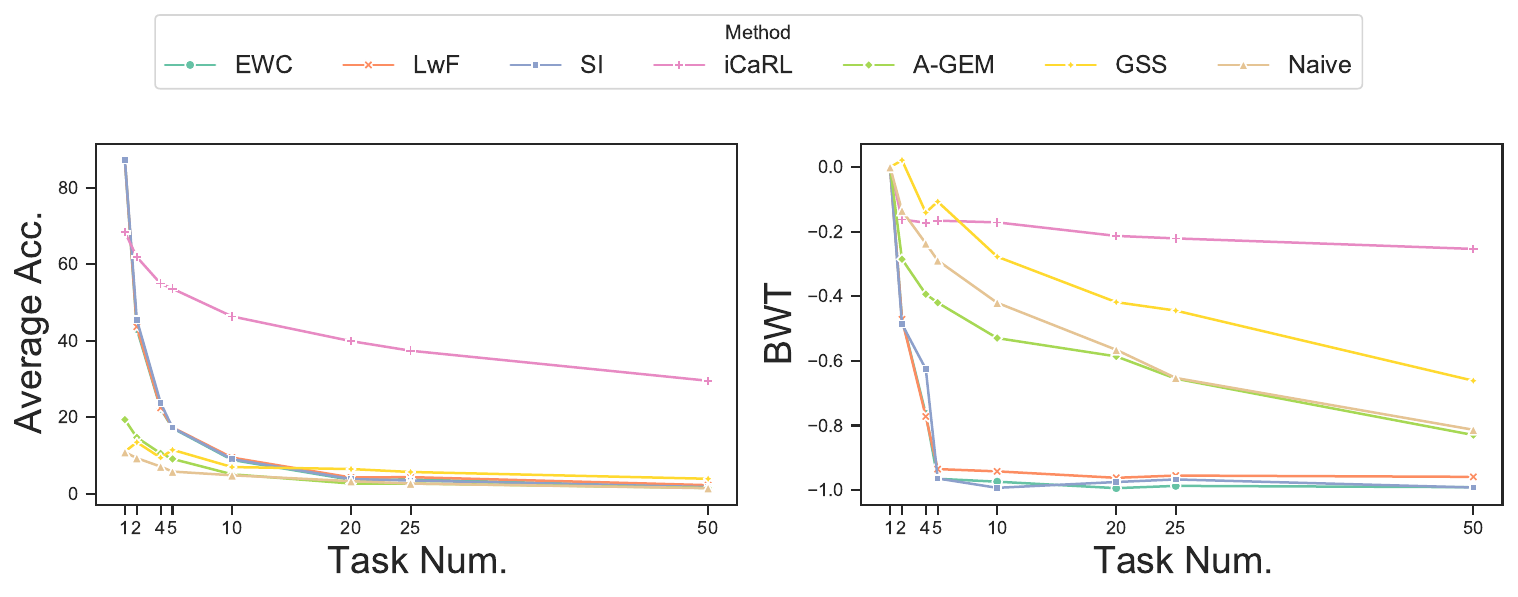}
\caption{Performance in varying lengths of task sequence on SplitCIFAR100. iCaRL shows an advantage in learning long task sequence while regularization-based methods are better at short ones. Specifically, When the number of tasks exceeds 10, all the methods except for iCaRL, only achieve model accuracy below 10\%. iCaRL can handle long task sequence to some extent with higher BWT than other methods, but its accuracy decreases to 30\% around when the length of task sequence reaches 50.}
\label{fig:scale}
\end{figure}

\subsection{Adaptability Analysis}

Adaptability is the fundamental model capacity for continual learning as it requires a model to continuously adapt to new tasks while preserving knowledge acquired from previous tasks. Therefore, to realize strong adaptability, a machine learning model is supposed to have high capacities of knowledge transfer and retention for learning new tasks without catastrophic forgetting. In addition, as the process of continual learning also accommodates uncertainty about when and where new tasks will arrive, a continual learning model is also expected to have strong adaptability in varying lengths of task sequences.

\textbf{Knowledge transfer and retention.} Fig.~\ref{fig:adapt} shows results of model capacities in knowledge transfer and retention on four benchmarks. 
Typically, an ideal continual learning model has strong adaptability both in knowledge transfer and retention. Thus, its average accuracy across tasks and backward transfer are supposed to be high as much as possible. However, achieving high average accuracy does not mean the model can learn each task successfully and high BWT solely also cannot reflect the model has realized knowledge retention. For example, in Fig.~\ref{fig:adapt}~(c), the BWT of iCaRL seems to remain stable especially when the length of task sequence varies from 5 to 10 or from 20 to 50. Nonetheless, its average accuracy on the task sequence of length 50 has dropped significantly compared to that on task sequence of length 20. Moreover, in Fig.~\ref{fig:adapt}~(c)-(d), GSS achieves higher BWT than other methods (except for iCaRL), but its average accuracy on each task sequence is not competitive or even lower. Therefore, simply measuring continual learning methods via any single metric is biased and cannot reflect their performance in a comprehensive and accurate way. Instead, a combination of evaluation metrics needs to be adopted for evaluating model capacities in different aspects.
Overall, methods evaluated have not achieved expected performance in knowledge transfer and retention, especially on large and difficult datasets.

\textbf{Varying lengths of task sequence.} Fig.~\ref{fig:scale} shows the potential of continual learning methods handling varying lengths of task sequence on SplitCIFAR100. Specifically, the capacity of knowledge transfer and retention of three regularization-based methods (LwF, EWC and SI) diminishes linearly as the number of tasks increases. They tend to perform like a sprinter during continual learning and are able to alleviate catastrophic forgetting to some extent when learning on a short task sequence. However, they usually fail on long task sequences especially after the number of tasks reaches 10, results of which are consistent with indications of previous studies\cite{hsu2018re,ahn2019uncertainty}.
Similarly, GSS and A-GEM shows poor performance especially in long task sequences.
By contrast, iCaRL shows stronger adaptability and tends to perform like an endurance athlete during continual learning. It has an advantage and the potential to continuously learn in a long task sequence, despite that its performance in learning short task sequences is not much significantly better.

\subsection{Sensitivity Analysis}

Having achieved strong adaptability, an ideal continual learning model is also supposed to be sensitive to dynamic task variations to deal with different types of differences between tasks and varying degrees of task granularity. Specifically, the different types of differences between tasks can be embodied in different continual learning scenarios\cite{Gido2019three,hsu2018re}, including Class-IL, Task-IL, and Domain-IL. The varying degrees of task granularity can be reflected in task sequence composed of fine-grained and coarse-grained tasks, where fine-grained tasks share more similarities between tasks while coarse-grained ones bear more differences.

\begin{figure}
\centering
 \includegraphics[scale=0.32]{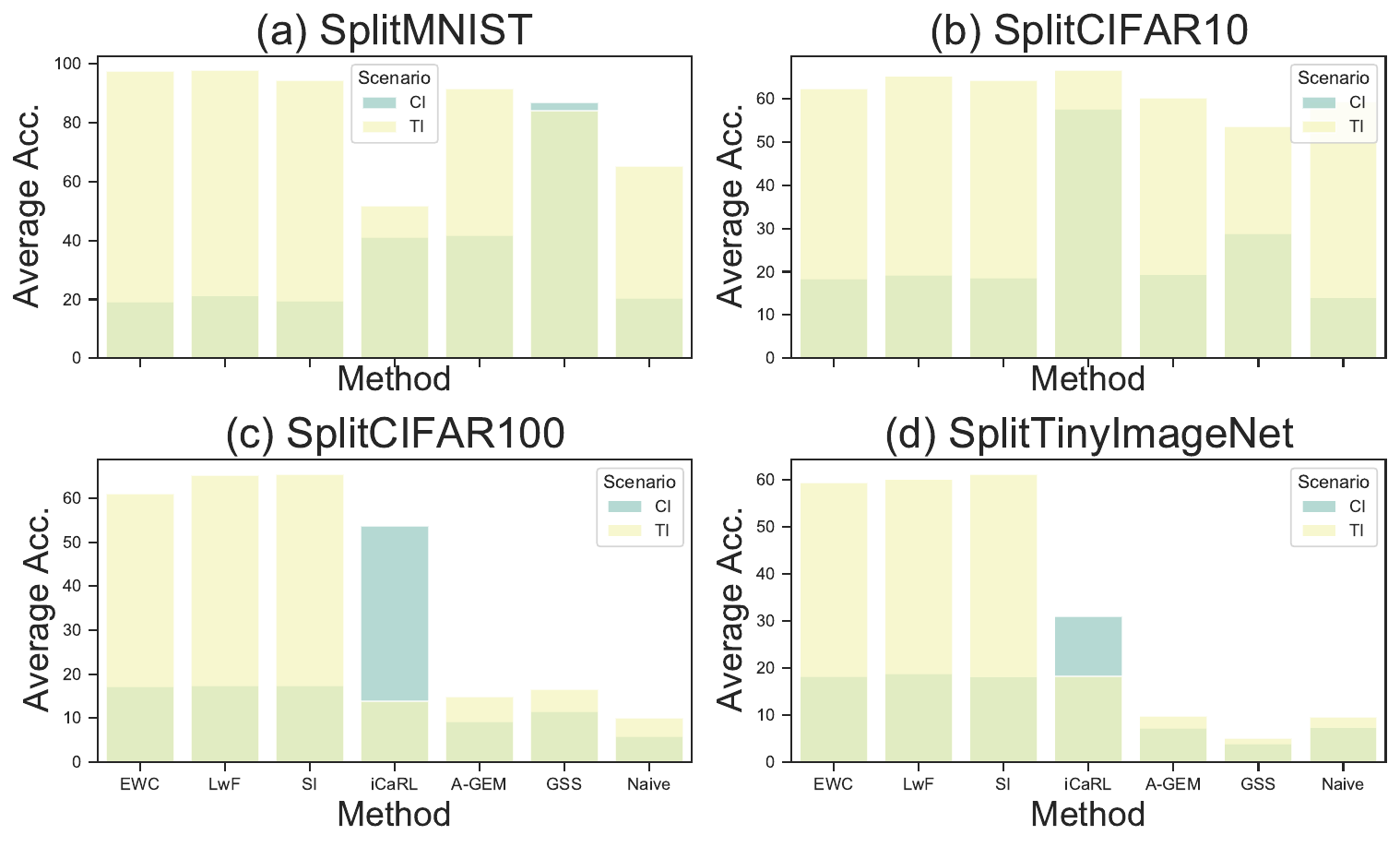}
\caption{Performance in various continual learning scenarios. ``CI'' and ``TI'' denotes Class-IL and Task-IL, respectively.
Most methods evaluated cannot handle various types of differences between tasks with significant performance drop in Class-IL. However, iCaRL performs better in Class-IL than Task-IL on SplitCIFAR100 and SplitTinyImageNet.}
\label{fig:scenario}
\end{figure}

\begin{figure}
\centering
\includegraphics[scale=0.32]{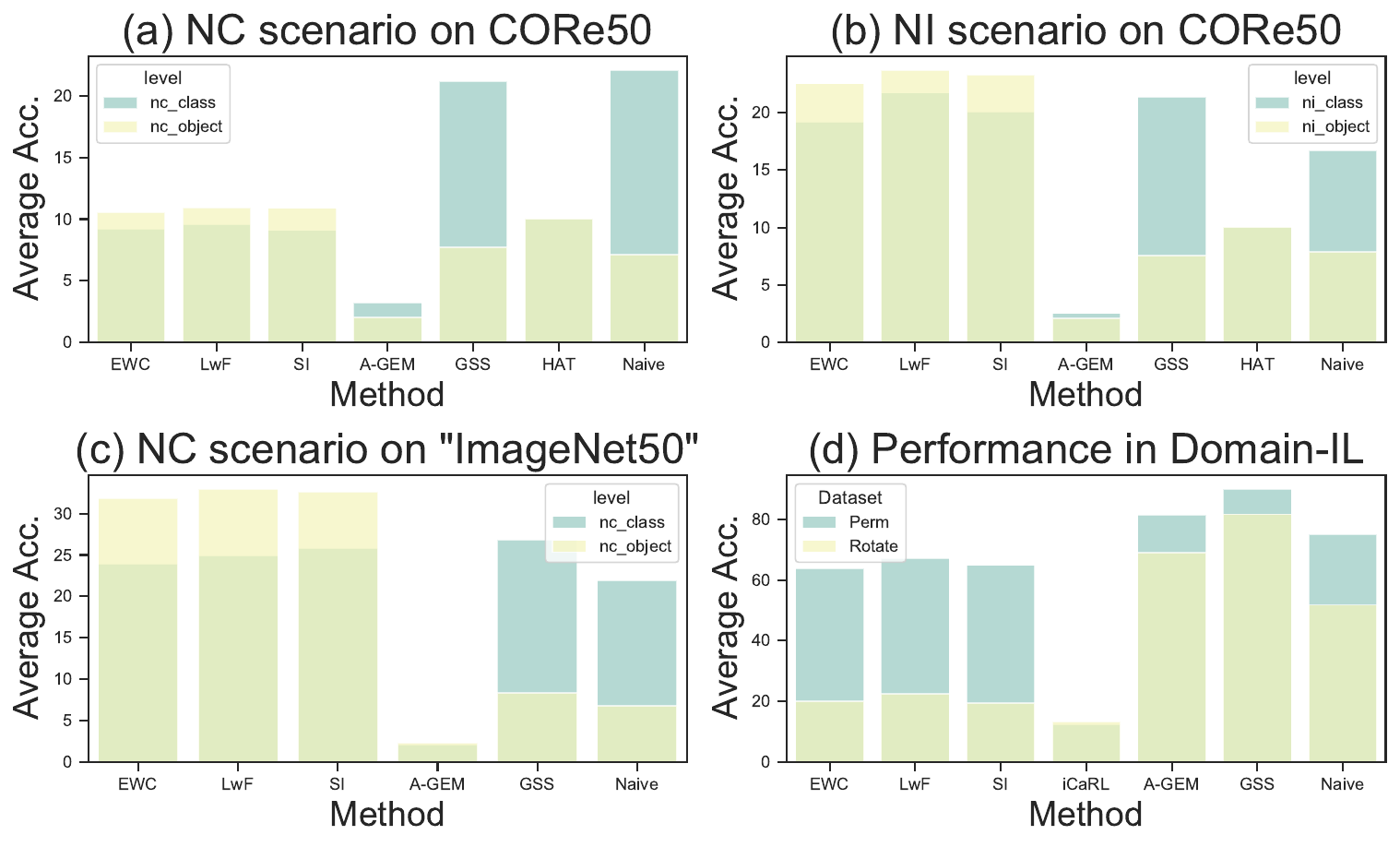}
\caption{Sensitivity results. (a)-(c) shows performance in varying task granularity and (d) shows performance in Domain-IL.
In (a)-(c), ``NC'' and ``NI'' refer to the New Classes and New Instances scenarios, respectively in\cite{lomonaco2017core50}. The ``class'' and ``object'' level denote multi-class classification at category level (10 categories) and object level (50 classes), respectively in CORe50.
Regularization-based methods are more inclined to fine-grained task sequence, while replay methods prefer coarse-grained tasks. HAT shows no preference for task relationships.
In (d), ``Perm'' and ``Rotate'' refer to PermutedMNIST and RotatedMNIST dataset, respectively.
Most methods evaluated (except for iCaRL) achieve better performance on PermutedMNIST than RotatedMNIST. iCaRL shows high capacity in difficult Class-IL but has the lowest accuracy in Domain-IL. GSS and A-GEM achieves higher performance in Domain-IL regardless of datasets.}
\label{fig:granu_di}
\end{figure}

\textbf{Different types of differences between tasks.}
Comparative results on three types of continual learning scenarios are shown in Fig.~\ref{fig:scenario} and Fig.~\ref{fig:granu_di}-(d). Generally,  methods evaluated have not been able to handle various task relationships embodied in different types of continual learning scenarios. For instance, although iCaRL can deal with the most difficult setting of Class-IL, it fails in Domain-IL. Moreover, most methods (except for iCaRL) prefer Task-IL where explicit task information is provided to help infer task identity. 
In addition, A-GEM and GSS perform well in Domain-IL.

\textbf{Varying task granularity.}
Results of model performance in varying task granularity are displayed in Fig.~\ref{fig:granu_di}: (a)-(d). Note that as iCaRL cannot be established on Core50 and ``ImageNet50'' using Avalanche\cite{lomonaco2021avalanche}, we exclude iCaRL in this experiment. 
From the results, it can be indicated that no method evaluated has achieved a trade-off in learning similarities and differences between tasks. Specifically, regularization-based methods perform better at object level than category level, suggesting these methods are more inclined to fine-grained tasks. Thus, regularization-based methods tend to focus on learning similarities between tasks. This result has validated explanations for the mechanism of regularization-based methods to alleviate catastrophic forgetting in existing studies\cite{GGerman2019Continual,van2020brain}.
By contrast, replay methods tend to focus on learning differences between tasks. For instance, GSS performs significantly better at category level than object level.
In addition, the dynamic architecture HAT achieves nearly the same performance at category and object level, showing that it has no preference for similarities or differences between tasks. This can be explained by that HAT dedicate separate network modules or parts to each task during continual learning.

To conclude, results show that methods we consider have divergent preferences to relationships between tasks and no method is able to handle both fine-grained and coarse-grained task sequences. 




\subsection{Efficiency Analysis}

Besides strong adaptability and sensitivity to dynamic task variations, a continual learning model is also expected to be efficient in both memory usage and training time consumption in practice. Specifically, we consider evaluating model efficiency in memory usage in terms of replay buffer size for replay methods, and evaluating model efficiency in training time consumption in terms of the number of training epochs. 

\begin{figure}
\centering
\includegraphics[scale=0.28]{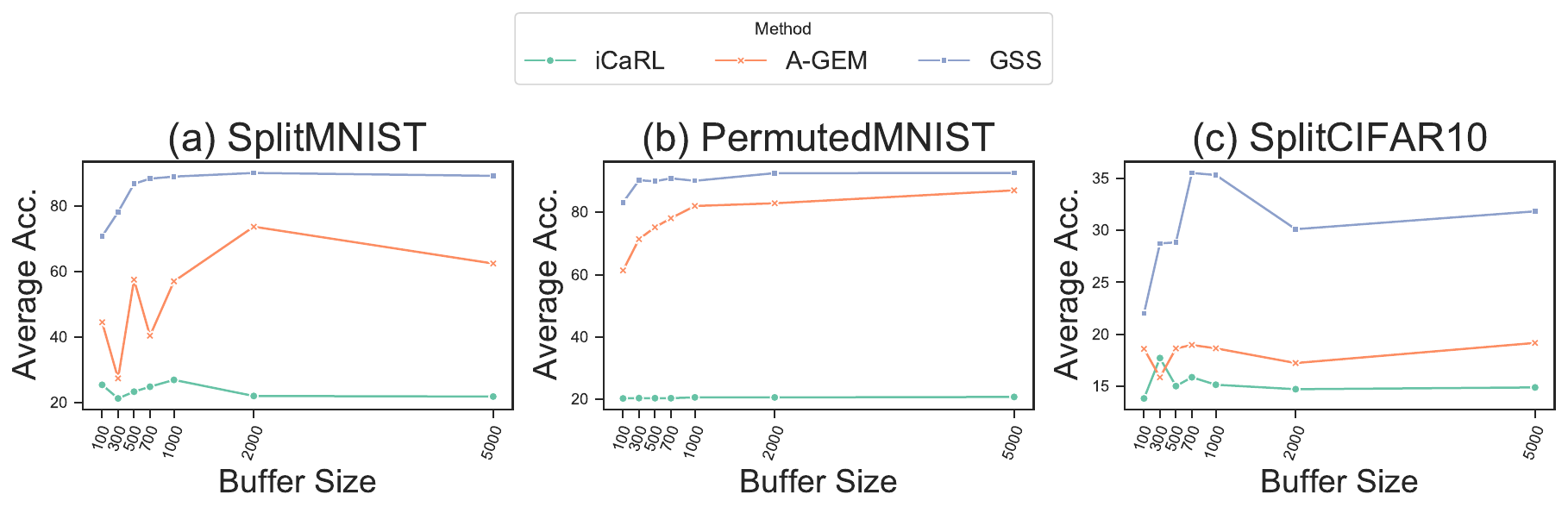}
\caption{Efficiency performance in varying sizes of replay (or memory) buffer. Given a task sequence, increasing buffer size does not ensure continuous improvement for replay methods, and there exits an optimal buffer size to make a balance between memory budgets, model performance and training time consumption. Results also indicates sample efficiency under the same buffer size.
}
\label{fig:buffer}
\end{figure}

\begin{figure}
\centering
\includegraphics[scale=0.28]{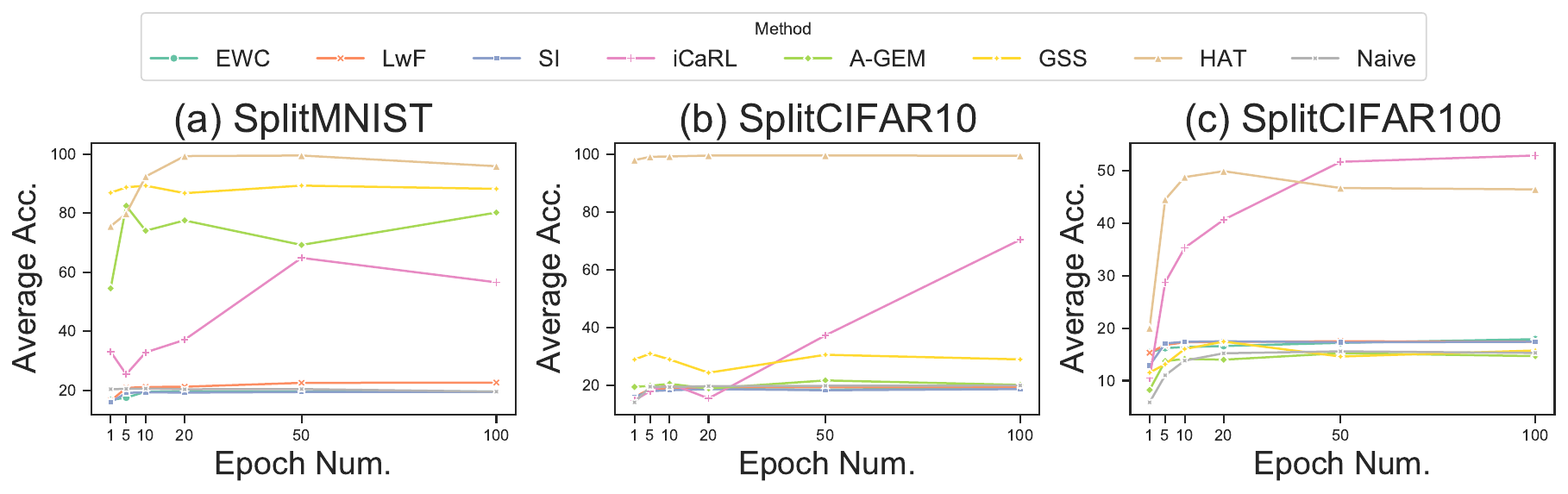}
\caption{Efficiency performance in varying numbers of training epochs. Given a task sequence, performance of most methods evaluated cannot be significantly improved by using many training epochs. This suggests that these methods (except for iCaRL) can be applied in online setting. Note that HAT learns with the help of task labels and cannot be directly compared to other methods.
}
\label{fig:epoch}
\end{figure}

\textbf{Efficiency in memory usage.}
Results of three replay methods with varying sizes of replay buffer are shown in Fig.~\ref{fig:buffer}. Experimental results demonstrate that there can be an optimal buffer size that makes a balance between model performance, memory budget for replay and training time consumption, given a certain task sequence during continual learning. In other words, when learning on a sequence of continuous tasks, methods using a replay buffer do not require the buffer size to continuously grow as the model capacity can be saturated after the buffer reaches a certain size. 
For example, iCaRL may not demand for a replay buffer in a very large size as shown in Fig.~\ref{fig:buffer}~(c). This is consistent with \cite{Matthias2021Defying} where iCaRL does not always benefit from an increasing replay buffer size for an easy-to-hard task ordering due to the reduced stochasticity in the estimated feature means in the nearest-mean-of-exemplars classifier. For A-GEM, its optimal buffer size should depend on the dataset and the optimal buffer size for GSS can be controlled below 1000.

Fig.~\ref{fig:buffer} also indicates sample efficiency of each replay method under the same buffer size. Results show that with a fixed size of replay buffer, GSS achieves the highest average accuracy. This means GSS stores and replays the most effective samples compared to A-GEM and iCaRL. 
Overall, results show that current methods with a replay buffer have achieved some degree of efficiency in memory usage, inspiring that improvement of replay methods should not simply relies on a large buffer size.

\textbf{Efficiency in training time consumption.}
Results of performance with varying numbers of training epochs are presented in Fig.~\ref{fig:epoch}.
As the measurement of training time requires tight variable control (i.e., CPU, GPU, CUDA, etc), we instead evaluate model efficiency in training time consumption via training epochs. Particularly, fewer training epochs mean less training time consumption. 
Results show that there seems no need of multiple training epochs for most of methods except for iCaRL. This means that these methods can be applied to a single-pass or online setting without sacrificing much performance. This result, however, is inconsistent with the training procedure in many existing studies, which assume that training in multiple epochs can significantly improve model performance for continual learning\cite{kirkpatrick2017overcoming,zenke2017continual,serra2018overcoming}.

Typically, when there is only one training epoch, it is the single-pass or online setting\cite{chaudhry2018efficient,aljundi2019gradient} in continual learning. This setting involves sample data provided one-by-one without the access to old data within the same task or from different tasks. The online setting satisfies continual learning assumption and conforms to real-world scenarios. Therefore, the efficiency of continual learning methods in the single-pass or online setting is an important property that should be considered in practice. Excluding HAT which is learned with the help of task labels, Fig.~\ref{fig:epoch}~(a)-(b) show that using only one training epoch, GSS achieves the best performance. The success may be due to the high sample efficiency of GSS as suggested in experiments with varying sizes of replay buffer. The sample efficiency thus may directly influence training efficiency and convergence speed of a continual learning model.

\section{Conclusion and Perspectives}
\label{future}
In this paper, we evaluate and analyze recent continual learning methods on a basis of desiderata proposed from perspectives of cognitive properties that support continual learning in human brains. The desiderata concern model capacities of adaptability, sensitivity and efficiency in continual learning. We further propose a systemic evaluation paradigm comprising evaluation protocols designed for each desideratum.

Experimental results across eight benchmarks demonstrate that although some methods exhibit some degree of adaptability and efficiency, no method we consider is able to identify task relationships when encountering dynamic task variations, or achieve a trade-off in learning similarities and differences between tasks. These methods also usually fail in long task sequence and their capacity of knowledge transfer and retention diminishes linearly with the increase of task sequence length, especially on large-scale and difficult datasets.
Notice that results reported in the paper can be influenced by some uncontrollable random factors (i.e., running times of an experiment, hardware and software environments, etc). Therefore,  we focus on the trends in experimental results for comparison.

Based on these findings, some guidance is provided for the improvement of continual learning models and their application in real-world scenarios. First, adaptability is the core and fundamental characteristic of continual learning, but none of existing methods has achieved strong self-adaptation. Therefore, future work needs to seek an adaptive solution to rapidly adapt to new task learning while maintaining knowledge and skills acquired from old tasks. 
The adaptive solution aims to learn in a dynamic environment involving continual sequential task learning and continuous dynamic changes. For example, learning multiple diverse tasks in game StarCraftII or achieving robust performance for a controller in automatic control systems require an adaptive solution to non-stationary environments.
Specifically, the adaptive solution is supposed to reach an equilibrium state under certain conditions despite that it may be a suboptimal solution to some individual tasks temporally. Thus, the formulation and computation of the adaptive solution can draw on the representation of and solutions to equilibrium state in cybernetics, evolution, game theory, etc.

Second, the identification of task relationships is an essential capacity for continual learning models, which is similar to the pattern separation and integration supporting human continual learning. Therefore, it is necessary to learn meta-information of a task, build an association model between tasks, and apply task relationships as a kind of knowledge during continual learning. Particularly, relationships between tasks can be modelled explicitly and learned in a self-supervised way. For example, a dynamic graph neural network (GNN) can be used to model task relationships. To satisfy the dynamics of continual learning where new tasks continuously emerge, the graph needs to be able to dynamically adjust its topology. Then task relationships can be extracted and propagate through the dynamic graph and learned as additional knowledge to be stored and leveraged by a continual learning model. 

Third, there is a need of making a balance between model efficiency and performance in continual learning. 
Although continual learning has not involved large-scale models, the number of tasks faced by continual learning models is inherently unknown. Usually, a complex neural network with deeper hidden layers is more efficient in representation learning to extract features from input. However, it is often difficult and time-consuming to train a large and complicated network. Therefore, future work is expected to achieve a trade-off between the effectiveness and simplicity of network architectures for continual learning models.
In addition, future work can investigate factors that may influence model efficiency besides the optimal replay buffer size and training epoch number studied in this paper. Finally, these factors can be tested as a reference for optimal hyperparameter settings, eliminating the need for hyperparameter tuning while ensuring model efficiency in continual learning.

In conclusion, current evaluation protocols cannot reflect whether a continual learning model has satisfied certain desiderata underlying their cognitive capacities. Thus, future work should not consider evaluating continual learning models simply via model accuracy, forgetting and other micro-level metrics. Instead, it demands for the desiderata proposed in this paper with corresponding evaluation protocols to examine model performance in different aspects. The evaluation paradigm based on cognitive properties is thus essential to help discover potential challenges and opportunities in continual learning. To summarize, this work is expected to provide inspirations and suggestions for establishing an ideal machine learning model that realizes truly continual learning.
\ifCLASSOPTIONcaptionsoff
  \newpage
\fi

\bibliographystyle{IEEEtran}
\bibliography{references}

\vspace{11pt}

\vspace{11pt}
\end{document}